\documentclass[conference]{IEEEtran}
\usepackage{ifsym}
\IEEEoverridecommandlockouts
\usepackage{cite}
\usepackage{amsmath,amssymb,amsfonts}
\usepackage{algorithmic}
\usepackage{graphicx}
\usepackage{textcomp}
\usepackage{multirow}
\usepackage{xcolor}
\def\BibTeX{{\rm B\kern-.05em{\sc i\kern-.025em b}\kern-.08em
    T\kern-.1667em\lower.7ex\hbox{E}\kern-.125emX}}
\begin{document}

\title{Object Isolated Attention for Consistent Story Visualization
}

\author{
    \IEEEauthorblockN{\textit{
        Xiangyang Luo$^{1,2}$, 
        Junhao Cheng$^3$,
        Yifan Xie$^1$, 
        Xin Zhang$^1$, 
        Tao Feng$^{1,2}$, 
        Zhou Liu$^1$, 
        Fei Ma$^1$\IEEEauthorrefmark{2}, 
        Fei Yu$^1$
    }\thanks{$^{\dag}$Corresponding author.}}
    \vspace{0.2cm}

    \IEEEauthorblockA{$^1$Guangdong Laboratory of Artificial Intelligence and Digital Economy (SZ), Shenzhen, China \\
    }
    \IEEEauthorblockA{$^2$Shenzhen International Graduate School, Tsinghua University, Shenzhen, China \\
    }
    \IEEEauthorblockA{$^3$Sun Yat-sen University, Shenzhen, China \\
    }
    \IEEEauthorblockA{
    \{goodluoxy, knowxyf, zhangx0526, fengtao0127\}@gmail.com, howe4884@outlook.com, 
    \{liuzhou, mafei, yufei\}@gml.ac.cn
    }
}
\maketitle
\begin{abstract}
Open-ended story visualization is a challenging task that involves generating coherent image sequences from a given storyline. One of the main difficulties is maintaining character consistency while creating natural and contextually fitting scenes—an area where many existing methods struggle. In this paper, we propose an enhanced Transformer module that uses separate self attention and cross attention mechanisms, leveraging prior knowledge from pre-trained diffusion models to ensure logical scene creation. The isolated self attention mechanism improves character consistency by refining attention maps to reduce focus on irrelevant areas and highlight key features of the same character. Meanwhile, the isolated cross attention mechanism independently processes each character’s features, avoiding feature fusion and further strengthening consistency. Notably, our method is training-free, allowing the continuous generation of new characters and storylines without re-tuning. Both qualitative and quantitative evaluations show that our approach outperforms current methods, demonstrating its effectiveness.
\end{abstract}


\section{Introduction}
Story visualization, the task of generating coherent image sequences from a narrative~\cite{storygan}, has emerged as a rapidly advancing field at the intersection of computer vision and natural language processing. This task holds immense potential for applications in education~\cite{education2, pointtalk}, entertainment~\cite{entertainment}, and beyond, where visual storytelling can greatly enhance narrative engagement and comprehension. However, generating high-quality, consistent visualizations that accurately reflect the storyline remains a significant challenge, especially when managing multiple characters and complex scenes.

Traditional story visualization methods have primarily relied on models trained on specific datasets~\cite{storytransformer,chen2022character}, limiting their generalization capabilities and real-world applicability. 
For instance, approaches like StoryGAN~\cite{storygan}, which employ generative adversarial networks~\cite{gan}, often struggle to maintain character identity and visual coherence throughout a narrative.
While recent advancements in diffusion-based models~\cite{makeastory,auto,storyimager,gong2023talecrafter} have improved image quality, they still face difficulties in generating new characters and scenes without extensive re-tuning.
With the rise of pre-trained text-to-image models~\cite{diffusion,latentdiffusion}, methods are now starting to achieve open-world comic generation, which holds much greater practical value. The current mainstream approaches to maintaining consistency can be broadly categorized into two types: (1) Specifying character positions with bounding boxes and using IP-Adapters~\cite{ye2023ip} to manipulate cross attention, aiming to preserve character identity. However, this method struggles to produce natural interactions and reasonable layouts~\cite{theatergen,autostudio,freecustom}. (2) Achieving character consistency through the concatenation of self attention mechanisms, which, despite its potential, has limited success in preserving character identity and often results in feature confusion among multiple characters~\cite{consistory,storydiffusion, dreamstory}.

Building on the concept of concatenated self attention~\cite{codeswap, jedi}, we propose a novel approach that alleviates feature confusion and enhances character consistency. We observe that different characters frequently reference each other, and images tend to pay insufficient attention to concatenated features. To address this, we introduce an isolated self attention mechanism~\cite{vaswani2017attention} that employs masks to prevent mutual attention between characters and enforces focus on the same character across sequences via cross attention information. Additionally, we design an isolated cross attention mechanism which leverages prior knowledge of reasonable layout composition in diffusion models and generates separate prompts for each character, further improving consistency and reducing feature mixing.
\begin{figure*}[t]
    \centering
    \includegraphics[width=\linewidth]{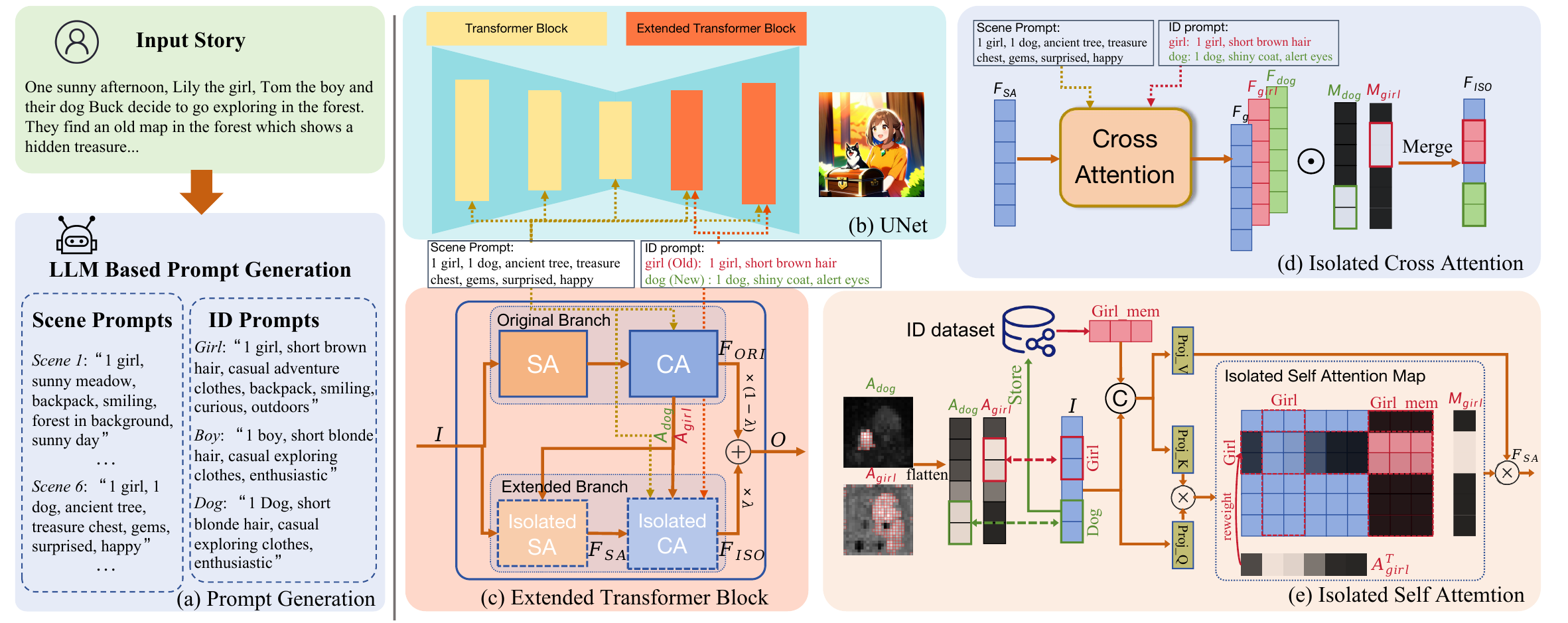}
    \caption{Pipeline of our framework. Given a story, we utilize an LLM agent to decompose it into scene prompts and character prompts (a). These prompts are then fed into a pre-trained diffusion model to generate anime-style images (b). We replace the traditional upsampling Transformer block with an extended Transformer block, which introduces an extended branch (c). In this new branch, we design isolated self attention (e) and isolated cross attention (d) mechanisms, which extract cross attention maps from the original branch to enhance character consistency and reduce feature confusion.}
    \label{fig:pipeline}
\end{figure*}
Our method operates in a training-free manner, allowing for the continuous generation of new characters and storylines without the need for re-tuning.
Our contributions can be summerized as follows:
\begin{itemize}
    \item We design an isolated self attention, which refines attention maps to reduce irrelevant focus and enhances attention on the specific reference character.
    \item We propose an isolated cross attention, which separates character features by utilizing the layout generated by diffusion models and individual character prompts.
    \item Extensive qualitative and quantitative experiments demonstrate the effectiveness of our approach in achieving visually consistent and coherent story visualizations.
\end{itemize}




\section{Related Work}

\subsection{Diffusion Models}
Diffusion models~\cite{ddpm} revolutionized text-to-image generation through iterative denoising processes. Subsequent work~\cite{LDM,sdxl,ramesh2022hierarchical} refined this approach using latent spaces to reduce computational costs while maintaining quality. Latent Diffusion Models (LDMs) emerged as a widely adopted framework for efficient synthesis across diverse generative tasks.
Diffusion Transformers (DiT)~\cite{dit} enhanced this paradigm by replacing U-Net with transformer-based designs, improving training efficiency and scalability. 
Despite varying structures, these models share the fundamental Transformer Block component, incorporating self-attention for image layout and cross-attention for conditional generation from other modalities. Our method modifies this block to achieve consistent story visualization without training, applicable to any existing diffusion framework.

\subsection{Story Visualization}


Story visualization~\cite{storygan} aims to generate coherent image sequences that align with multi-sentence paragraphs, enabling dynamic storytelling. Traditional methods~\cite{storytransformer,chen2022character} often rely on training models on specific datasets annotated with scene or character-level details. While these methods achieve consistency within limited domains, their scalability is constrained by data dependence and lack of flexibility.

The emergence of text-to-image diffusion models has opened new avenues for tuning-free story generation. These approaches can be broadly categorized into two directions. The first direction integrates pre-trained models~\cite{theatergen, autostudio} like IP-Adapter~\cite{ye2023ip} with bounding box inputs, enabling multi-character control. 
However, such methods often struggle to model complex character interactions and suffer from visual artifacts, such as the 'copy paste' effect. The second direction leverages self-attention mechanisms~\cite{storydiffusion, consistory, freecustom} to concatenate features across sequences, which is also applied to wide field such as video and 3D
 generation~\cite{instant3d, grid, human}. While this approach reduces the need for explicit annotations, it faces challenges such as feature leakage and inconsistent outputs, particularly in modeling long-range dependencies across storylines.

Our work adopts the latter approach and addresses these limitations by isolating the attention mechanism to prevent feature leakage and enhance consistency. This design ensures more robust alignment between textual descriptions and generated visual sequences, enabling more natural and coherent story visualization compared to existing methods.

\section{Method}
The workflow of our method is illustrated in Fig.\ref{fig:pipeline}. Starting with a given storyline, we utilize a Large Language Model (LLM)~\cite{openai2024chatgpt} to generate scene prompts $S = \{S_{1}, S_{2}, \dots, S_{N}\}$ and character prompts $C = \{C_{1}, C_{2}, \dots, C_{K}\}$. For each scene \( S_i \), we define three sets of indices: \( \mathcal{I}_i \), \( \mathcal{I}_i^{\text{new}} \), and \( \mathcal{I}_i^{\text{old}} \), representing the indices of all characters present in the scene, characters appearing for the first time, and characters reappearing from previous scenes, respectively. The image generated for scene $i$ is expressed as $I_{i} = \Theta(z, S, C^{i})$, where $z$ represents the initial noise, $C^{i}$ denotes the character prompts for scene $i$ (obtained from $C$ using $\mathcal{I}_i$), and $\Theta$ represents our extended diffusion model. As shown in Fig.~\ref{fig:pipeline} (b) and Fig.~\ref{fig:pipeline}  (c), our model enhances the traditional diffusion process by replacing the standard Transformer block with our extended Transformer block during the up-sampling stages. This extended Transformer block consists of two branches: the original branch, identical to the traditional Transformer block, and an extended branch. The extended branch incorporates cross-attention maps from the original branch and combines isolated attention mechanisms to improve character consistency across the generated images. The process of generating masks from the cross attention map is detailed in Sec.~\ref{sec:maskgen}, while the isolated self attention and cross attention mechanisms are introduced in Sec.~\ref{sec:self} and Sec.~\ref{sec:cross}, respectively.

\subsection{LLM Based Prompt Generation}
We designed a comprehensive pipeline to transform user-provided storylines into fully developed comic series utilizing an LLM agent, whose input and output are shown in Fig.~\ref{fig:pipeline}(a). The process begins with storyline refinement, where the LLM enhances the initial simple narrative by expanding plot elements, developing subplots, and enriching character backgrounds to create a detailed story manuscript. This manuscript is then segmented into distinct scenes to form a structured storyboard, ensuring logical flow and pacing. For each identified scene, the LLM generates detailed scene prompts $S$ that outline the visual and contextual elements necessary for illustration. Concurrently, the system creates individual character prompts $C$ that include physical descriptions, personality traits, and unique attributes to maintain consistent character depiction throughout the comic. Finally, a character-scene mapping is established, indicating which characters appear in each scene and if it is first time to appear, thereby facilitating coordinated and accurate visual representation. This sequential and integrated pipeline ensures a seamless transition from a basic storyline to an engaging and visually coherent comic series.

\subsection{Mask Generation}
\label{sec:maskgen}
To implement our subject isolation attention, we first need to capture the positions in the image before generating it. The cross-attention in the original branch provides an approximate location of the subject. Therefore, we use the Otsu method~\cite{otsu} to segment the subject from the image, formulated as: 
\begin{equation}
    M_m = Otsu(C_m),
\end{equation} 
where $C_m$ represents the cross-attention map of character $m$, and $M_m$ is the corresponding mask. To address noise in the cross-attention map, we average it with the attention map obtained from the previous denoising steps.

However, we empirically observe that the pre-trained diffusion model assigns varying levels of attention to different words, and the noise levels in attention maps also vary. Typically, attention maps with higher noise tend to segment irrelevant regions. To address this, we compute the coefficient of variation for each attention map, which with lower coefficients tend to produce more accurate segmentations. When attention maps overlap, we prioritize maps with lower coefficients of variation for better accuracy.

\subsection{Isolated Self Attention}
\label{sec:self}

Based on the obtained masks, we propose Isolated Self Attention to enhance the objects' consistency, which is illustrated in Fig.~\ref{fig:pipeline} (e). For each item in $\mathcal{I}_i^{\text{new}}$, we store the relevant tokens as references for use in subsequent scenes. This process is defined as $F_{m}=I \odot M_{m},$ where $m \in \mathcal{I}_i^{\text{new}}$ represents the index of the new character, $F_m$ is the set of selected tokens, $n_m$ is the number of tokens in $F_m$, and $\odot$ denotes the Hadamard product. Irrelevant tokens are excluded to prevent feature fusion in the later steps. For each item in $\mathcal{I}_i^{\text{old}}$, the relevant tokens are retrieved for reference in the following attention calculations. The queries $Q$, keys $K$, values $V$, and the vanilla attention map $A$ are formulated as:
\begin{align}
    &Q = \phi_Q(I), \nonumber \\
    &K = \phi_K\left[\mathrm{Concat}\left(I, \{F_{m} \mid m \in \mathcal{I}_i^{\text{old}}\}\right)\right], \nonumber \\
    &V = \phi_V\left[\mathrm{Concat}\left(I, \{F_{m} \mid m \in \mathcal{I}_i^{\text{old}}\}\right)\right], \nonumber \\
    &A = Q \times K,
\end{align}
where $\phi_Q$, $\phi_K$, and $\phi_V$ are three distinct linear transformations and $A \in \mathbb{R}^{(h \ast w) \times (h \ast w + \sum_m n_m)}$ for $\forall m \in \mathcal{I}_i^{\text{old}}$. Here, $h$ and $w$ represent the height and width of the image latent, respectively, and $n_m$ is the number of tokens in $F_m$.

Although we select tokens from the character regions as references, each region in the generated image still attends to all previous characters, which leads to feature confusion and reduces consistency. To address this, we refine the vanilla attention mechanism with two components: an attention mask that ensures each subject attends only to itself, thus preventing feature confusion, and attention re-weighting, which increases focus on the reference subject to further enhance consistency.

\subsubsection{Attention Mask}
\label{sec:mask}
The attention mask ensures that regions unrelated to a specific subject do not attend to that subject, thereby preventing feature confusion. We identify the reference tokens corresponding to each subject by locating the indices within the range from $s$ to $s + n_m$, where $\mathbf{s}$ is the starting column index of the corresponding character. A mask is then applied to ensure that only the region associated with the current subject can attend to these reference tokens, formulated as:
\begin{equation}
A[: , s:s + n_m] \mathrel{=} M_m
\end{equation}
for each $m \in \mathcal{I}_i^{\text{old}}$. For simplicity, we omit the broadcasting mechanism that aligns the shapes of the two tensors. This process is illustrated in Fig.~\ref{fig:pipeline} (e), ensuring that each character only attends to its corresponding tokens, thereby improving consistency and reducing feature confusion.

\subsubsection{Reference Re-weight}

\begin{figure}[h]
    \centering
    \includegraphics[width=\linewidth]{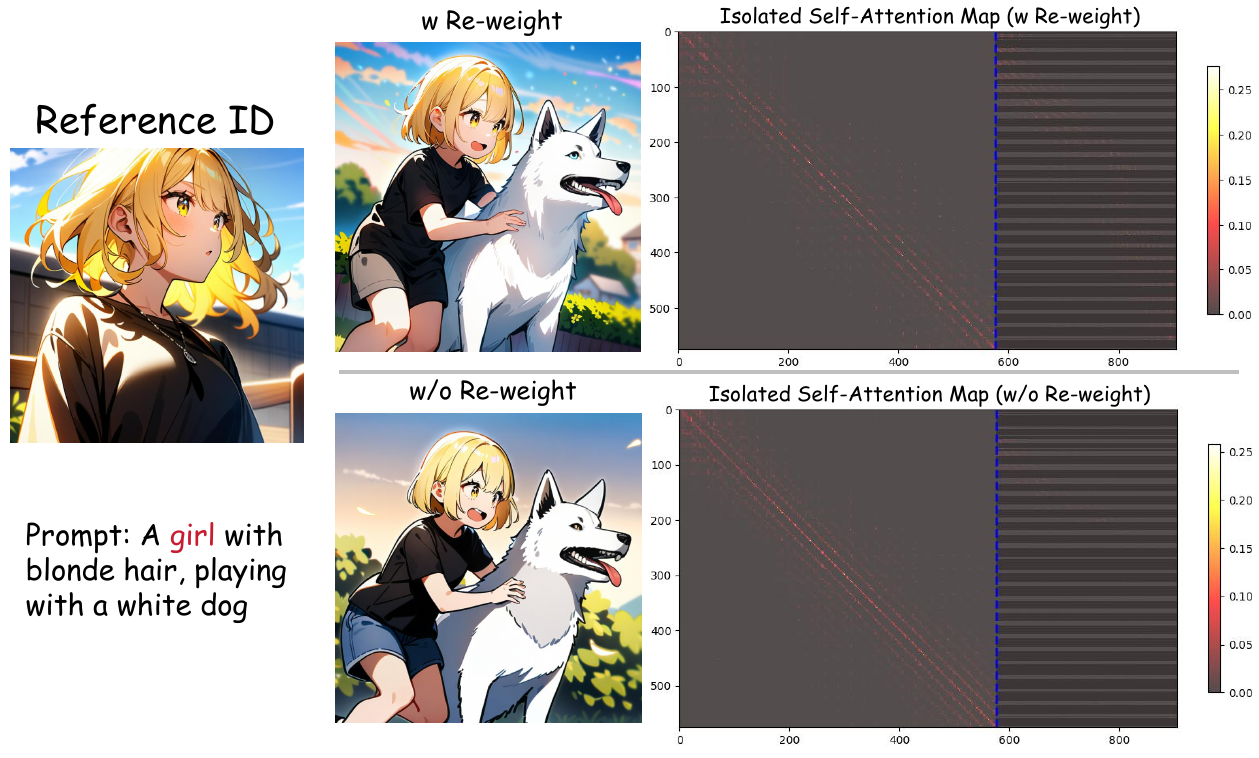}
    \caption{Ablation study of our re-weight operation and the visualization of the isolated self attention map, which reveals that after re-weight, the character's skin tone, hair color, and image style align more closely with the reference image. The attention map also shows increased focus on the reference tokens, with non-gray areas indicating regions masked by the operation described in Sec.~\ref{sec:mask}.}
    \label{fig:reweight}
\end{figure}

HD-Painter~\cite{hdpainter} discovers that during inpainting, the modified region tends to focus excessively on the surrounding areas, diminishing the effectiveness of the inpainting process. Similarly, through our visualizations of self attention, we observe that tokens in the character regions disperse too much attention to surrounding areas, resulting in insufficient focus on the reference tokens, as shown in Fig.~\ref{fig:reweight}, which leads to reduced consistency. To mitigate the influence of surrounding areas on the character regions, we re-weight the self attention map based on the activation level of each token in the cross attention map relative to the specified prompt.

Specifically, for each old character $m$, we normalize the corresponding cross attention map as follows:
\begin{equation}
C_m = \text{Clip}\left(\frac{C_m - \text{median}(C_m)}{\text{max}(C_m)}, 0, 1\right)
\end{equation}
where Clip is a clipping operation between $[0,1]$, max$(\cdot)$ refers to the operation of finding the maximum value, and median$(\cdot)$ refers to the operation of finding the median value. We then re-weight the extended self attention map using the obtained $C_m$ and the mask $M_m$. The re-weighting process is shown in Fig.~\ref{fig:pipeline} (c) and can be expressed as:
\begin{equation}
A[M_m, :h \times w] = \mathrm{Rep}(C_m, n_m, 1), \quad \forall m \in \mathcal{I}_i^{\text{old}},
\end{equation}
where the repeat operator $\text{Rep}(\cdot, n_m, 1)$ replicates a column vector $n_m$ times into a matrix. As demonstrated in Fig.~\ref{fig:reweight}, our reference re-weighting technique ensures better alignment between the character's skin tone, hair color, and overall image style with the reference image, highlighting the effectiveness of our re-weight operation.

\subsection{Isolated Cross Attention}
\label{sec:cross}

Due to the limited ability of text-to-image models to fully comprehend complex prompts, generating scenes with multiple characters often results in feature confusion, where descriptions intended for one character are incorrectly applied to another. To tackle this issue, we adopt the concept of regional prompts~\cite{mixofshow, migc, layout}, which allows for the independent update of character-specific features using unique prompt words for each character. In the final step, these character-specific features are blended with global features using a mask. However, unlike typical approaches that rely on externally defined bounding boxes, we utilize the binary masks $M$ derived in Sec.~\ref{sec:mask}, following the inherent layout composition of the diffusion model. This method results in more natural-looking images. The process is formulated as:
\begin{equation}
    F_{\text{ISO}} = \left(1 - \bigcup_{m \in \mathcal{I}_i^{\text{old}}} M_m\right) \odot F_{\text{CA}}^{\text{g}} + \sum_{m \in \mathcal{I}_i^{\text{old}}}M_m \odot F_{\text{CA}}^m,
\end{equation}
where $F_{\text{CA}}^{\text{g}}$ and $F_{\text{CA}}^m$ represent the output features of cross attention with the scene prompt and ID prompts, respectively, as illustrated in Fig.~\ref{fig:pipeline} (d). As shown in Fig.~\ref{fig:cross}, when there are many character descriptions, traditional methods often suffer from feature confusion, whereas our method effectively captures the prompts and generates accurate images.
\begin{figure}[h]
    \centering
    \includegraphics[width=\linewidth]{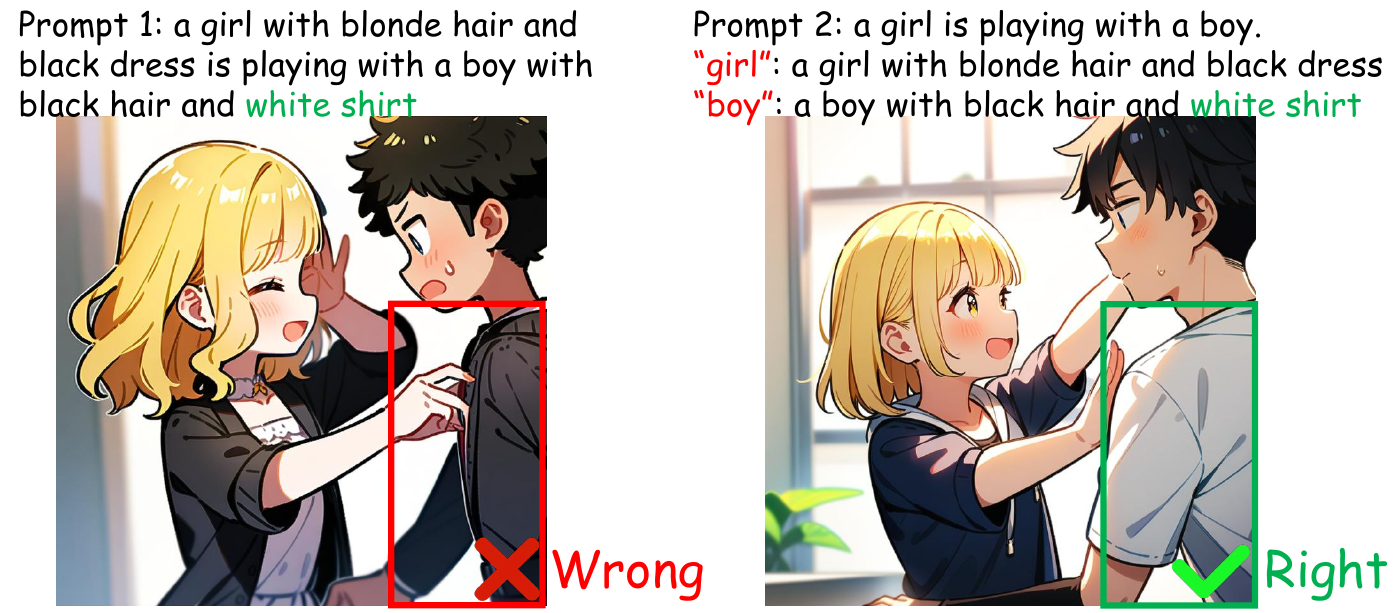}
    \caption{Comparison with common cross attention with our isolated cross attention. Our method accurately isolates the character’s features, preventing confusion between black and white clothing.}
    \label{fig:cross}
\end{figure}

\subsection{Branch Merge}
As shown in Fig.~\ref{fig:pipeline} (c), we obtain two output features, $O_{ori}$ and $O_{ISO}$, which are derived from the original branch and the extended branch, respectively. The original branch outputs the basic features, while the extended branch generates features with stronger attribute consistency by utilizing our isolation attention mechanism. These two features are linearly weighted to produce the final feature output of this block:
\begin{equation}
    F = F_{ISO} \times \lambda + F_{ORI} \times (1 - \lambda),
\end{equation}
where the lambda is a hyperparameter. Inspired by the concept of Classifier-Free Guidance~\cite{cfg}, we set $\lambda$ to a value greater than 1, allowing the output features to deviate from the original features to achieve stronger consistency results. Based on empirical results, the optimal value for $\lambda$ is 1.1.
\section{Experiments}
In this section, we compare our method with some existing open-ended story visualization methods to show our effectiveness. The experiments settings are introduced in Sec.~\ref{sec:setting}, the quantitative and qualitative comparison are shown in Sec.~\ref{sec:quantitative} and Sec.~\ref{qualitative} respectively.

\subsection{Experimental Settings}
\label{sec:setting}
We utilize the test dataset from \cite{theatergen}, focusing specifically on the multi-turn story generation component. This dataset, which was both automatically generated and manually curated, consists of 4,000 stories, each comprising four scenes. Some prompts in the dataset containes pronouns like \textquotedblleft they" to refer to characters, so we rewrite these prompts using ChatGPT to clarify the descriptions and reduce ambiguity. We compare our method against several existing approaches, including StoryDiffusion~\cite{storydiffusion}, Intelligent Grimm~\cite{intelligent}, Mini-DALLE3~\cite{minidalle}, and ChatGPT4o~\cite{openai2024chatgpt}.
Our method is built on SDXL~\cite{sdxl}, which is the same as StoryDiffusion~\cite{storydiffusion}, and it is noticed that our method can be employed in all existing diffusion models.
\subsection{Quantitative Results}
\label{sec:quantitative}
To validate the effectiveness of our approach, we adopt four evaluation metrics. For overall image quality, we utilize CLIP~\cite{clip} to calculate the Text-Image Similarity (TIS), which measures how accurately the generated images reflect the text. Additionally, we employ an aesthetics predictor~\cite{schuhmann2024improved} to assess the aesthetic quality (AQ) of the images, reflecting their visual appeal. 
To evaluate character consistency, we utilize Grounding-DINO~\cite{groundingdino} to detect the target subjects based on class tokens. Using the first occurrence of the subject as a reference, we calculate the CLIP similarity (IIS)~\cite{clip} and DreamSim similarity (DS)~\cite{dreamsim} between subsequent occurrences of the same subject and the reference. The quantitative results, shown in TABLE~\ref{tab:comparison}, demonstrate that our method outperforms the other approaches across all metrics.
\begin{table}[h]
\centering
\renewcommand{\arraystretch}{1.3} 
\scriptsize 
\caption{Quantitative comparison of existing methods with ours.}
\begin{tabular}{l|c|c|c|c}
\hline
\multirow{2}{*}{\textbf{Method}} & \multicolumn{2}{c|}{\textbf{Comprehensive Metrics}} & \multicolumn{2}{c}{\textbf{Subject Consistency}} \\ \cline{2-5} 
 & \textbf{TIS} $\uparrow$ & \textbf{AQ} $\uparrow$ & \textbf{IIS (\%)} $\uparrow$ & \textbf{DS (\%)} $\uparrow$ \\ \hline
Mini-DALLE3~\cite{minidalle} & 21.91 & \underline{6.47} & 57.34 & 34.31 \\ \hline
Intelligent Grimm~\cite{intelligent} & 24.96 & 5.14 & 61.82 & 32.99 \\ \hline
ChatGPT4o~\cite{openai2024chatgpt} & 27.38 & 6.11 & 58.16 & 45.83 \\ \hline
StoryDiffusion~\cite{storydiffusion} & \underline{28.39} & 6.42 & \underline{68.75} & \underline{46.56} \\ \hline
Ours  & \textbf{28.91} & \textbf{6.52} & \textbf{70.28} & \textbf{49.63} \\ \hline
\end{tabular}
\label{tab:comparison}
\end{table}

\subsection{Qualitative Result}
\label{qualitative}

The qualitative result is illustrated in Fig.~\ref{fig:cross}. Given a sequence of prompts, our method can accurately capture the prompt to generate the correct content, and the character consistency between different prompts surpasses others. Other methods experienced feature fusion, leading to a decline in image quality and character missing. 
ChatGPT4o~\cite{openai2024chatgpt} accurately interprets the text due to its powerful base model, but it still falls short in maintaining character consistency through different images, even when style and consistency are emphasized in the given prompts.

\begin{figure}[h]
    \centering
    \renewcommand{\arraystretch}{1.1}
    \includegraphics[width=\linewidth]{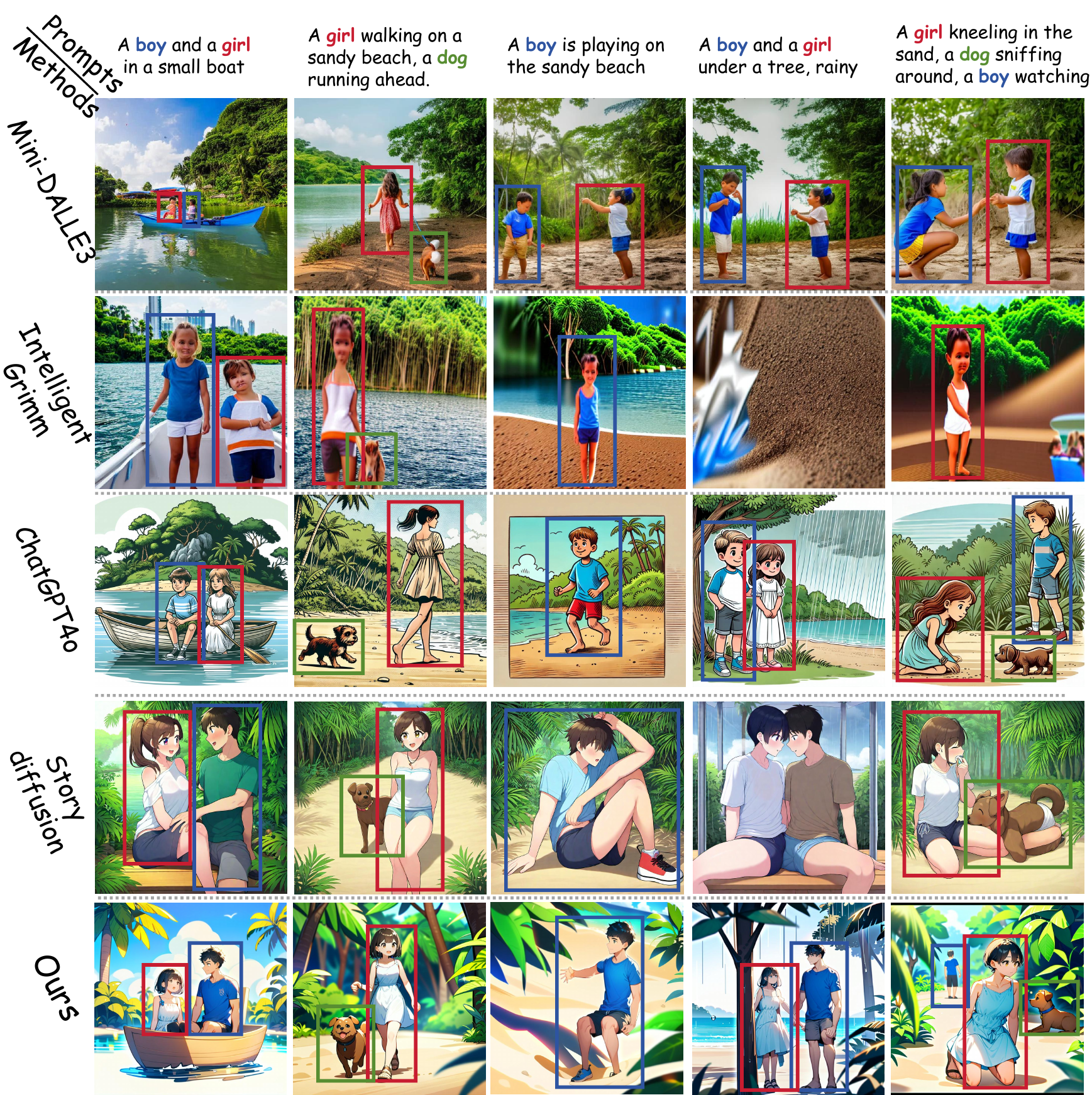}
    \caption{Qualitative comparison results. Each column of images should match the content of the prompt, and the appearance of characters within all the same-colored bounding boxes in each row should remain consistent. The results demonstrate that our method effectively maintains character consistency and accurately aligns with the prompt content.}
    \label{fig:cross}
\end{figure}
\subsection{Ablation Study}
To validate effectiveness of our isolated self attention and isolated cross attention,  we conduct ablation experiments on our components. 
We have three components: IC denotes the isolated cross-attention, IS denotes the isolated self-attention, and Re denotes the reweight of self-attention, which should cooperate with IS. 
When none of the components are present, this metric reflects the capability of the base model. The results are shown in TABLE~\ref{tab:results}. 

For TIS and AQ, since our goal is solely to improve consistency, which can not improve these metrics and even may harm them, but our method maintains them at comparable levels. In contrast, StoryDiffusion, which uses the same base model, shows a significant drop in TIS and AQ compared to the base model, further highlighting the effectiveness of our approach in avoiding such degradation. For IIS and DS, which reflect consistency, each module of our method achieves notable improvements independently. The results shows the effectivenss of our proposed modules.


\begin{table}[h]
    \centering
    \renewcommand{\arraystretch}{1.1}
    \caption{Quantitative results of ablation study.}
    \begin{tabular}{|c|c|c|c|c|c|c|c}
        \hline
        \multirow{2}{*}{IC} & \multirow{2}{*}{IS} & \multirow{2}{*}{Re} & \multicolumn{2}{c|}{\textbf{Comprehensive Metrics}} & \multicolumn{2}{c|}{\textbf{Subject Consistency}} \\ \cline{4-7}
        & & &\textbf{TIS} $\uparrow$ & \textbf{AQ} $\uparrow$ & \textbf{IIS (\%)} $\uparrow$ & \textbf{DS (\%)} $\uparrow$ \\
        \hline
        $\times$ & $\times$ & $\times$ & 28.93 & 6.55 & 64.87 & 44.10 \\ 
        $\checkmark$ & $\times$ & $\times$ & 28.74 & 6.46 & 67.31 & 46.39 \\ 
        $\times$ & $\checkmark$ & $\times$ & 29.11 & 6.55 & 68.89 & 48.36 \\ 
        $\times$ & $\checkmark$ & $\checkmark$ & 28.96 & 6.54 & 69.12 & 48.95 \\ 
        $\checkmark$ & $\checkmark$ & $\checkmark$ & 28.91 & 6.52 & 70.28 & 49.63 \\
        \hline
        \multicolumn{3}{|c|}{StoryDiffusion~\cite{storydiffusion}} & 28.39 & 6.42 & 68.75 & 46.56 \\
        \hline
    \end{tabular}

    \label{tab:results}
\end{table}


\section{Conclusion}
In this paper, we introduce isolated attention mechanisms to enhance character consistency and prevent feature confusion in story visualization. Specifically, we retain the prior layout of the image generated by diffusion models, capture positional information from the cross attention map, and introduce isolated self attention and isolated cross attention mechanisms to prevent unnecessary information fusion and enhance the focus on relevant information. 
Experimental results show that our approach outperforms existing methods in maintaining character identity and generating coherent visual narratives. 
This work proposes a new approach to consistency generation and could be considered for applications in consistent video generation and 3D video generation.

\bibliographystyle{IEEEtran}
\bibliography{IEEEabrv, OIA}


\end{document}